\documentclass{article}

\PassOptionsToPackage{numbers, compress}{natbib}   % numeric citations, e.g. [1, 3--5]
\usepackage[dblblindworkshop, final]{neurips_2026}
\workshoptitle{Medical Reasoning with Vision Language Foundation Models (Med-Reasoner)}   % REQUIRED for the workshop track: appears at the foot of page 1 in [final] mode

\usepackage[utf8]{inputenc}
\usepackage[T1]{fontenc}
\usepackage{hyperref}
\usepackage{url}
\usepackage{booktabs}
\usepackage{amsfonts}
\usepackage{amsmath}
\usepackage{nicefrac}
\usepackage{microtype}
\usepackage{xcolor}
\usepackage{graphicx}
\usepackage{multirow}
\usepackage{tabularx}
\usepackage{ragged2e}
\usepackage{makecell}
\usepackage{array}


\title{A Deterministic Evidence Layer for Vision-Language Autism Screening from Naturalistic Home Video}

\author{%
  Wenqi Li$^{1,\dagger}$ \quad Mindi Ruan$^{2,\dagger}$ \quad Chuanbo Hu$^{1}$ \quad Shuo Wang$^{3,*}$ \quad Xin Li$^{1,*}$ \\[4pt]
  $^{1}$Department of Computer Science, University at Albany, SUNY, Albany, NY, USA \\
  $^{2}$Lane Department of Computer Science and Electrical Engineering, \\
  West Virginia University, Morgantown, WV, USA \\
  $^{3}$Department of Radiology, Washington University in St.\ Louis, St.\ Louis, MO, USA \\[2pt]
  $^{\dagger}$Co-first authors. \quad $^{*}$Corresponding authors: \texttt{shuowang@wustl.edu},\ \texttt{xli48@albany.edu}
}

\begin{document}

\maketitle

\begin{abstract}
Autism spectrum disorder (ASD) is diagnosed through specialist observation of a child's social behavior, and access to that expertise is the bottleneck for early identification. Vision-language models (VLMs) describe a child's behavior from video well; the verdict drawn from the description is unstable: at temperature~0, across eight pipeline configurations on one backbone, 16--37\% of clips change their predicted label between repeated runs, and the cause lies in the serving stack. We keep the VLM frozen and move the decision out of the model. Under our grounded perception constraints the VLM writes an event table of timestamped, glossary-labeled events that names the eliciting press and logs counter-evidence; a text-only stage age-calibrates the confidence of each row; a deterministic weight-of-evidence scorer sums it into an evidence total and stratifies it into a risk category, so every decision decomposes into named per-feature contributions and can be re-scored from the saved table. On 43 caregiver-recorded, protocol-free home free-play clips of preschool children, the pipeline reaches AUC $0.851 \pm 0.012$, 86.0\% accuracy, and F$_1$ 71.8 over three runs. It labels 74.4\% of clips correctly in every run (60.5\% for the zero-shot baseline) and flags no typically developing clip in every run (9 of 31 at zero-shot). An ablation on the same backbone attributes the gain to the grounded perception constraints read through the deterministic scorer.
\end{abstract}

% CLAIM GATES (carried over from the Frontiers manuscript, v0.9):
% (1) The 0.851 / 86.0 / 94.6 headline lives on its own axis (Gemini 3.1 Pro + score.py risk category).
%     NEVER present it as a delta over the Gemini 2.5 Pro single-call baseline (73.0 five-run mean) ---
%     different backbone AND decision rule; the matched-axis comparison was removed (2026-09-06), so no cross-backbone deltas anywhere.
% (2) "Perception is the causal root cause" of the temp-0 flips stays OFF the table until a
%     >=10-seed variance decomposition separating Stage 1 from Stage 2 is run (Limitations / Future Directions).

%%%%%%%%%%%%%%%%%%%%%%%%%%%%%%% Introduction %%%%%%%%%%%%%%%%%%%%%%%%%%%%%%%
\section{Introduction}
\label{sec:intro}

Autism spectrum disorder (ASD) is diagnosed by watching a child: a specialist observes social communication and repetitive behavior, usually with a standardized instrument such as the Autism Diagnostic Observation Schedule, Second Edition (ADOS-2)~\cite{lord2012ados2}, against DSM-5 criteria~\cite{apa2013dsm5}. That expertise is scarce, waits for evaluation are long~\cite{gordonlipkin2016wait}, and the lifetime costs of ASD are substantial~\cite{buescher2014economic}. Home video is the signal families can supply without a clinician: a phone, a few minutes of play, no protocol. We ask whether a vision-language model (VLM) can turn such video into a screening decision worth acting on, positioned after a positive caregiver-report screen~\cite{robins2014mchatr} as a triage step during the wait for evaluation.

A VLM can describe what a child does in a clip, and the description is often good. The decision drawn from it is not reliable: when we run screening pipelines built on Gemini repeatedly at temperature~0, 16--37\% of clips change their label between runs across eight pipeline configurations on one backbone, five repeats of an identical request spread accuracy by 7.0 points, and the mechanism lies in the serving stack~\cite{thinkingmachines2025nondeterminism}, out of reach of anyone building on a closed API. A screening decision that changes between identical runs is not a clinical tool, and a decision that is the same every time is useful only if it is also right. The quantity we target is therefore stable correctness: the share of clips a pipeline labels correctly in every run.

\begin{figure}[t]
    \centering
    \includegraphics[width=\textwidth]{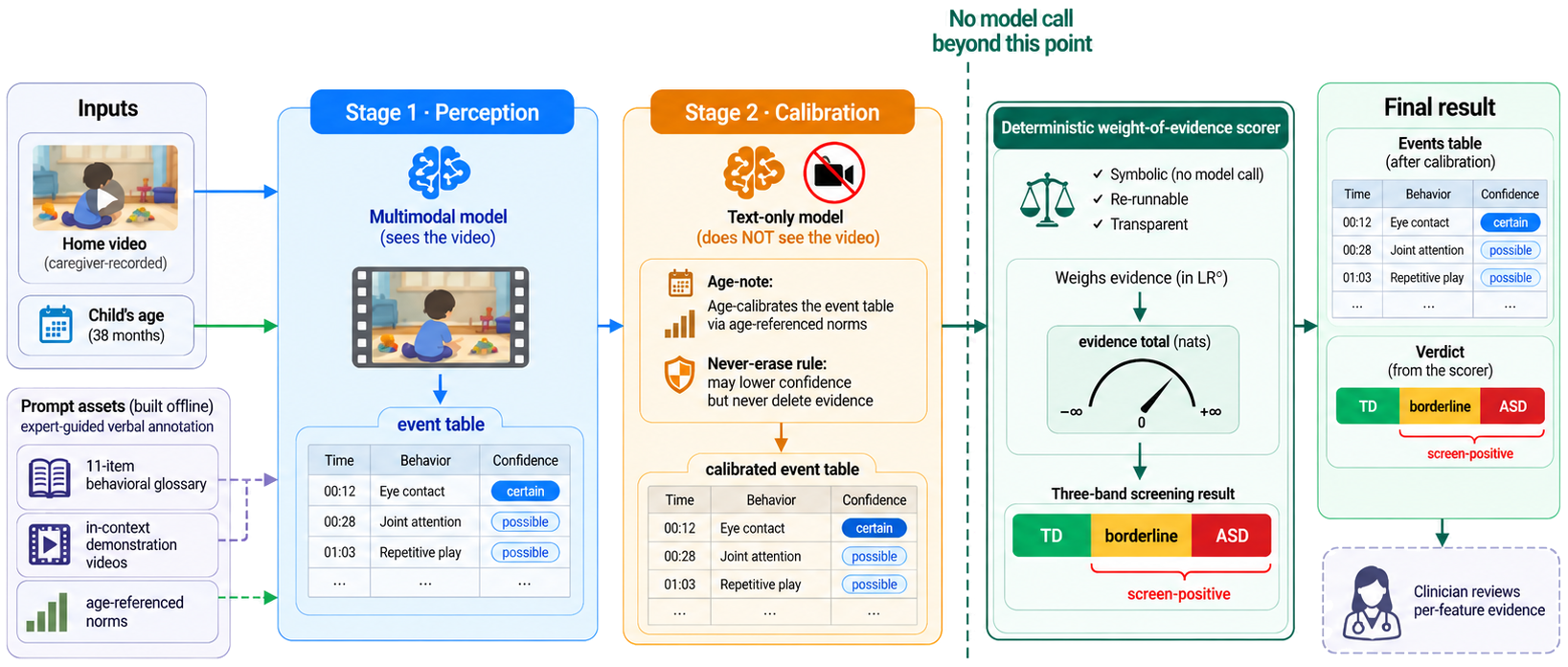}
    \caption{The screening pipeline. Stage~1, a VLM that sees the video, writes a timestamped event table under grounded perception constraints; Stage~2, a text-only model given the table and the child's age, calibrates each row under the never-erase rule; the deterministic scorer sums the calibrated table into an evidence total and stratifies it into a low-, intermediate-, or high-risk category.}
    \label{fig:method_overview}
\end{figure}
Two responses dominate the emerging literature. One line of work audits frozen general-purpose models on fine-grained clinical behavior tasks and finds that they handle coarse tasks such as activity recognition but fail on fine-grained ones such as activity segmentation and anxiety detection~\cite{kommineni2024childinclusive,deng2024avasd}; another trains reliability in, fine-tuning the backbone on clinician-rated videos~\cite{honarmand2026finetunemllm}.
We take a third approach (Fig.~\ref{fig:method_overview}): keep the model frozen and take the verdict away from it. In our pipeline the VLM is used only as a high-recall perception front-end. Prompted with a clinically aligned behavioral glossary, it must commit to timestamped, event-level observations (including explicit counter-evidence) that a human can check against the video. A text-only second stage then age-calibrates each event under a \emph{never-erase} rule: it can lower the weight of an observation and cannot delete it. A deterministic $\ln\mathrm{LR}^{+}$ weight-of-evidence scorer sums the calibrated table into an evidence total and stratifies it into a risk category. The diagnostic reasoning thus splits into a perceptual step checked against the video and a probabilistic step checked by arithmetic. Every label decomposes into named per-feature contributions, can be re-scored without a model call, and exposes a tunable recall--specificity operating point for triage. Our main contributions are:
\begin{enumerate}
    \item \textbf{An auditable, training-free VLM screening architecture that separates grounded observation from a deterministic, evidence-based decision.} A frozen VLM writes a timestamped, glossary-labeled event table; a text-only stage age-calibrates it; a deterministic weight-of-evidence scorer turns it into an evidence total and a risk category, so every decision decomposes into named per-feature contributions and can be re-scored without a model call. On 43 caregiver-recorded, protocol-free free-play clips it reaches AUC 0.851, and its decision rule is two explicit cut-offs at interpretable evidence levels, the social-communication cap and the weight of a single certain stereotypy (Appendix~\ref{subsec:scorer_twostage}).
    \item \textbf{Grounded perception constraints for VLM reasoning.} Stage~1 must commit to events with rationales confined to the interval; social-response deficits are coded only as failed opportunities with the eliciting press named; typical behaviors are logged as explicit counter-evidence. In the cumulative ablation the constraints lift specificity from 91.4\% to 94.6\% and, read through the scorer designed for these tables, raise F$_1$ from 50.7 to 71.8 and accuracy from 77.5\% to 86.0\%.
    \item \textbf{A stable-correctness account of VLM-based screening.} At temperature~0, 16--37\% of clips change label between repeated runs. The full pipeline labels 74.4\% of clips correctly in every one of three runs, against 60.5\% for the zero-shot baseline; no typically developing (TD) clip is flagged in every run (9 of 31 at zero-shot), three ASD clips are missed in every run, and the remaining variation enters through the event table.
\end{enumerate}

%%%%%%%%%%%%%%%%%%%%%%%%%%%%%%% Related Work %%%%%%%%%%%%%%%%%%%%%%%%%%%%%%%
\section{Related Work}
\label{sec:related}

\paragraph{Screening ASD from video.}
Classical pipelines quantify engineered behavioral signals with hand-crafted features and shallow statistical analyses or classifiers: gaze measured by eye tracking \cite{klin2002visual,jones2013eyes} or estimated from video \cite{yu2024socialgaze_arxiv}, facial dynamics \cite{guha2018facialdynamics}, head and body pose \cite{hashemi2012computer}, and speech acoustics \cite{hu2025speechpatterns}. Learned image and video models, deep \cite{ruan2021firstperson,ruan2023mmsys_decisiontree} or few-shot over handcrafted facial-dynamics descriptors \cite{zhang2023fewshot_taffc}, raise accuracy on protocol-elicited recordings, where micro-expression analysis is already limited by video quality \cite{ruan2024microexpressions}; most models were built under clinical protocols and degrade in unconstrained home scenes \cite{debelen2020cvreview}. Caregiver-recorded home video has long been recognized as a scalable screening substrate \cite{abbas2018jamia,babu2024nejmai,tariq2018mobilehomevideo}, and the strongest current results are supervised: ensembles over pose, speech, and object features from protocolized one-minute recordings reach AUROC 0.83 (0.73 external) on 510 children \cite{kim2025homevideoscreening}. Two recent neighbors bring VLMs to this substrate: zero-shot Gemini feature scoring on 50 public YouTube clips fed into pre-validated logistic classifiers \cite{azizian2025mllmhomevideo}, and LoRA fine-tuning of Gemini on 400 clinician-rated home videos \cite{honarmand2026finetunemllm}. We differ from the supervised line and both neighbors on three axes at once: no elicitation protocol, no training anywhere in the pipeline (including the decision layer), and a private evaluation corpus.

\paragraph{Decomposed and verifiable medical reasoning.}
Separating perception from downstream reasoning is by now a recognized design pattern in both general multimodal systems \cite{rapid2026decoupling} and clinical applications: text-only diagnostic pipelines fuse LLM-extracted evidence with deterministic rules \cite{depllm2026interview}, LLMs have been applied to autism-associated language disorders \cite{hu2025npjllmasd}, and a two-stage VLM analysis of parent--child interaction video finds expert alignment strong at the observation layer but weak at the judgment layer \cite{chi2026parentchild}, supporting a separation between observation and judgment. A growing line of work likewise reserves the final say for a layer other than the LLM: Prolog-based clinical decision support \cite{procds2025prolog}, LLMs demoted to explainers of statistical models \cite{diagnollm2025explainer}, and deterministic aggregation over expert-validated ECG observations \cite{zeta2025ecg}; probabilistic combination of evidence for diagnosis goes back to the earliest proposals for computer-aided diagnosis~\cite{ledley1959reasoning}, and its additive form, a sum of log-likelihood-ratio weights, to early clinical decision-support systems~\cite{spiegelhalter1984weights}. Radiology reasoning models instead enforce evidence-diagnosis consistency through reward shaping during training \cite{hero2026radiologyrl}. A parallel line adapts or trains vision-language models for the clinical domain~\cite{moor2023gmai,li2023llavamed}; here the backbone stays frozen. We apply this separation to raw, unprotocolized home video. Prior systems either transcribe clinician-led assessment recordings into behavioral scripts and let an LLM issue the verdict from the script~\cite{liu2024script}, or train a video classifier of stereotyped behaviors with auxiliary text supervision on public clips~\cite{deng2024languageassisted}; we contribute a different interface object: a probe-gated, timestamped event table with explicit counter-evidence under a never-erase rule, consumed by a likelihood-ratio scorer, so the label is consistent with the logged evidence by construction.

\paragraph{Reliability of clinical (V)LM decisions.}
Run-to-run instability of LLM outputs persists at temperature~0; its mechanism lies in batch-variant inference kernels and is not fixable from the user side of a closed API~\cite{thinkingmachines2025nondeterminism}. Clinical text measurements are accumulating: a repeatability framework over 912{,}000 generations~\cite{shyr2025llmrepeatability} and management-recommendation flips across five identical queries of the same vignette~\cite{jgim2026flipvariation}; the remedies these studies report are prompt design, re-prompting, and sampling or ensembling several models, with the serving-stack fix left to providers~\cite{thinkingmachines2025nondeterminism}. Agent benchmarks measure consistency across repeated attempts with pass$^{k}$~\cite{yao2024taubench}. We measure clip-level label flips across repeated runs of a clinical video pipeline and report pass$^{3}$ (Section~\ref{subsubsec:stability}), and the remedy is architectural: the decision rule is a fixed function of the saved event table and adds no variance of its own, so the variance that remains enters through the table, where it can be measured.

\section{Method}
\label{sec:method}

The pipeline is training-free, and its label is not produced by a model. A frozen VLM supplies perception; a deterministic evidence layer, made of a behavioral glossary, a structured event schema, and a weight-of-evidence scorer, holds the clinical knowledge and the decision (Fig.~\ref{fig:method_overview}). A clip passes through three steps: 1) Stage~1, one call to the VLM, turns the video into a table of timestamped, glossary-labeled events under grounded perception constraints; 2) Stage~2, one call to a text-only model, re-weights each row against the child's age without deleting anything; 3) the scorer sums the table into an evidence total and stratifies it by two fixed cut-offs into a risk category. Every clip yields three outputs: the calibrated event table, the evidence total, and the risk category. Because the label is arithmetic on a saved table, any run can be re-scored without a model call, and a screen-positive decision traces to specific rows.

\subsection{Clinical Cohort and Home-Video Corpus}
\label{sec:dataset}

The cohort consists of preschool-aged children who completed comprehensive autism evaluations at an academic medical center; participant-level ASD or typically-developing (TD) status was determined by licensed clinicians using ADOS-2~\cite{lord2012ados2} and DSM-5 criteria~\cite{apa2013dsm5} with consensus review. Evaluation labels for this study are clip-level ASD/TD ratings confirmed by independent ASD specialists. From a broader home-recording protocol we analyze only the \emph{free-play} segment of 47 children; four clips serve as in-context demonstrations, leaving a 43-clip evaluation set (12 ASD, 31 TD; per-group ages in Table~\ref{stab:cohort}). Caregivers filmed one continuous 3--6 minute segment of the child at play in the home, on their own smartphones or tablets, with no standardized prompts, scripted interactions, or elicitation presses. Nothing is required of families beyond a camera, and the footage carries spontaneous social, communicative, and sensorimotor behavior.

Expert knowledge enters through \emph{expert-guided verbal annotation}: a domain expert thinks aloud over the video, recording start/end timestamps, a behavior category, an atypical/typical judgment, and an observable-cue rationale (roughly twice real time per clip). These traces ground an 11-item glossary mapping ADOS-2 Social Affect and Restricted/Repetitive Behavior constructs and DSM-5 criteria to observable descriptors (Table~\ref{tab:glossary}, Appendix~\ref{app:glossary}), and supply the gold-answer exemplar tables of Section~\ref{subsec:knowledge}.
All procedures were approved by the institutional review boards of the participating institutions; parents provided written informed consent covering recording, research use, and AI-based analysis. Direct identifiers were removed and each recording was assigned a study code; the recordings are handled as identifiable human-subjects data under the approved protocol. The recordings remain under the originating institutions' data-use protocols and are not publicly distributed; no synthetic data are used anywhere in this work.

\subsection{Problem Formulation}

Let $v$ denote a short home video clip captured during free play, $a$ the child's age in months, and $y \in \{\text{ASD}, \text{TD}\}$ the clip-level expert rating. We construct a screening function $f: (v, a) \mapsto (\mathcal{E}', s, \hat{y})$ that returns a calibrated event table $\mathcal{E}'$, an evidence total $s$, and a risk category $\hat{y}$. Instead of training a video classifier, we decompose screening into two fixed model calls, which we call stages, followed by a deterministic decision layer:
\[
\begin{gathered}
    M_{\text{percept}}(v;\Pi_1) \rightarrow \mathcal{E} = \{e_k\}_{k=1}^{K},
    \qquad
    M_{\text{judge}}(\mathcal{E}, a;\Pi_2) \rightarrow \mathcal{E}', \\[3pt]
    s = \textsc{score}(\mathcal{E}') = \textstyle\sum_{f} w_f c_f,
    \quad
    \hat{y} = \mathrm{stratify}(s) \in \{\text{low}, \text{intermediate}, \text{high}\},
    \quad w_f = \ln \mathrm{LR}^{+}_f,
\end{gathered}
\]
where the perception model $M_{\text{percept}}$ is a VLM that sees the video, the judgment model $M_{\text{judge}}$ is text-only and never sees $v$, $c_f$ is a confidence factor, the sum is subject to the domain cap of Section~\ref{subsec:scorer}, and $\mathrm{stratify}$ assigns a risk category by two fixed cut-offs, with both \emph{intermediate} and \emph{high} risk screen-positive. Each event is a tuple $e_k = (t_k^{\text{start}}, t_k^{\text{end}}, g_k, r_k, q_k)$: a temporal extent, a label $g_k$ from the behavioral glossary $\mathcal{G}$ (eleven ASD-related features plus a typical-behavior marker), a short observable-evidence rationale $r_k$, and a confidence $q_k \in \{\texttt{certain}, \texttt{possible}\}$.

\subsection{Stage~1: Perception into a Grounded Event Table}
\label{subsec:twostage}
\label{subsec:reasoning_scaffold}
\label{subsec:knowledge}
\label{subsec:implementation}

Clinical observation separates what was seen from what it means: a rater first logs behaviors, then interprets them. Stage~1 is the logging step, a single call to the frozen VLM. Its input is the video; the child's age is withheld, so that observation is not shaped by expectation. Its prompt has four parts, each with one job (Appendix~\ref{app:prompts}): (a) the behavioral glossary (Table~\ref{tab:glossary}, Appendix~\ref{app:glossary}), which fixes the label space and defines each feature in observable terms; (b) four demonstration videos, each paired with its expert event table (Section~\ref{sec:dataset}), which show the output format and the labeling conventions, delivered multi-turn, each video as its own user turn and its table as the assistant turn, so that every table is bound to its video; (c) a four-step scaffold that orders the work into intermediate steps, in the spirit of chain-of-thought prompting~\cite{wei2022cot}: 1) propose candidate time intervals; 2) describe the observable behavior in each (mental-state attribution is discouraged); 3) assign a glossary label ($g_k \in \mathcal{G}$) or mark the behavior as typical; 4) write a rationale restricted to cues visible inside the interval; (d) the grounded perception constraints below, which state when a feature may be coded.

The output is an event table with one row per observed behavior and six columns: Start and End (timestamps), Behavior (a short observable description), Comment (the matched glossary feature, or a typical-behavior marker), Note (context such as the adult action that preceded the behavior), and Confidence (\texttt{certain} when clearly visible; \texttt{possible} when occluded, brief, or marginal). Typical behaviors (clear eye contact, responsive smiling, pointing to share, reciprocal turn-taking, pretend play) are logged as their own rows, explicit counter-evidence, so the table records what argued against a deficit as well as for it. Later steps parse the table programmatically.

\textbf{Grounded perception constraints.} Video VLMs hallucinate temporal content~\cite{vidhalluc2025cvpr} and lose accuracy once an answer must be tied to a time in the clip~\cite{videozerobench2026}, so a rationale may cite only cues inside its own interval. The six social-communication features name behaviors that are reduced or absent, and an absence can be written down without any occasion to show the behavior; the press rule and the deferral of \#4 therefore count an absence only against an occasion the table records, as ADOS-2 does with its presses~\cite{lord2012ados2}. Feature numbers below refer to the glossary (Table~\ref{tab:glossary}, Appendix~\ref{app:glossary}). How a feature may be coded depends on the evidence it needs, and the eleven features fall into three classes: (a) \emph{local features} (\#1 eye contact, \#2 facial expression, and \#6--\#11, from lack of pretend play to unusual sensory responses) are coded whenever observed in an interval; (b) \emph{probe-gated response features} (\#3 reduced gestures and joint attention, \#5 impaired reciprocity and response to name) are coded only when an adult makes a social bid, a \emph{press} such as calling the child's name, pointing at an object, or holding out a toy, and the child fails to respond; absence without a press is not coded, the press is named in the Note column (e.g., \texttt{name called $\times$2 $\rightarrow$ no orient}), and a \#3 or \#5 row without one is invalid; (c) the \emph{pervasive feature} (\#4 limited peer interest and quality of social overtures) is not coded in Stage~1, because it is a judgment about the whole clip, so Stage~1 logs the raw social material (overtures and shared bids as typical-behavior rows, sustained solitary or parallel play as its own row) and leaves the decision to Stage~2.

\subsection{Stage~2: Text-Only Age Calibration Under the Never-Erase Rule}
\label{subsec:judgment}

Stage~2 is the interpreting step, a second call to a text-only model. Its input has four parts, the Stage~1 table, the child's age in months, age-referenced behavioral norms, and two hand-written demonstration exchanges of the \#4 rule delivered as prior turns (Appendix~\ref{app:prompt_stage2}); it never sees the video. Most glossary features name a departure from an age expectation: pretend play is not expected before 18 months, solitary or parallel play is age-typical to about 42 months, and echoing of words is part of language learning to about 30 months (norms in Appendix~\ref{app:norms}); ADOS-2 selects its module and algorithm by age and language level~\cite{lord2012ados2}, and DSM-5 states that the manifestations of ASD vary with developmental level and chronological age~\cite{apa2013dsm5}. The cohort spans 26 to 48 months, across the ADOS-2 Toddler/Module~1 boundary at 30 months (Table~\ref{stab:cohort}), so a table read without age could not exempt the youngest children's age-typical behavior.

\looseness=-1 Stage~2 performs three operations in order: 1) it synthesizes the pervasive feature \#4 from the whole table, checking counter-evidence first, so that any clear overture, approach, or shared bid anywhere in the table blocks the row, and it is added only for a diffuse, consistent lack of social interest with adults available yet no overtures, with the weighed counter-evidence stated in the Note; 2) it calibrates each remaining row against the norms for the child's exact age, lowering Confidence from \texttt{certain} to \texttt{possible} and writing an age note in the Note column when the behavior is common at that age (repeated questioning judged age-appropriate, for instance), and keeping \texttt{certain} when the deficit clearly exceeds age expectation in intensity, duration, or pervasiveness; 3) it re-emits the full table. Under the \emph{never-erase rule} the only permitted changes are the lowered confidence, the age note, and the added \#4 row: Stage~2 may not delete a row, raise a confidence, or relabel a coded feature as typical. The asymmetry is deliberate. Every row remains a visually grounded observation, and Stage~2's reading, which is not grounded, can lower its weight and cannot remove it. The output is the calibrated event table; the verdict Stage~2 also writes is used only as the LLM verdict of the ablation (Section~\ref{subsec:ladder}).

\subsection{Deterministic Weight-of-Evidence Scorer}
\label{subsec:scorer}

The scorer reads the calibrated table and produces the evidence total and the risk category in four steps. 1) Each glossary feature $f$ carries a weight equal to its log positive-likelihood-ratio~\cite{jaeschke1994lr} (Good's weight of evidence~\cite{good1950weighing}),
\[
    w_f = \ln \mathrm{LR}^{+}_f = \ln\!\frac{\mathrm{sens}_f}{1 - \mathrm{spec}_f},
\]
with provisional per-feature recall (sensitivity) and specificity set from published effect sizes and prevalence estimates (Appendix~\ref{app:scorer_theory}); summing weights follows the Bayesian weight-of-evidence interpretation (derivation and design rationale in the same appendix). 2) Each feature counts \emph{once}, at its highest observed confidence ($c_f = 1.0$ for \texttt{certain}, $0.25$ for age-downgraded \texttt{possible}), so repeated events cannot inflate the score. 3) Because social-communication (SC) features are positively correlated, their sum saturates at the SC cap $\kappa_{\mathrm{SC}} = 1.5$ nats, while the near-independent restricted/repetitive (RRB) features sum in full:
\[
    s = \min\!\Bigl(\textstyle\sum_{f \in \text{SC}} w_f c_f,\; \kappa_{\mathrm{SC}}\Bigr) + \sum_{f \in \text{RRB}} w_f c_f .
\]
4) Two fixed cut-offs stratify $s$ into three risk categories, low ($s<1.5$), intermediate ($1.5 \le s < 1.8$), and high ($s \geq 1.8$), with intermediate and high both screen-positive. The cut-offs sit at interpretable evidence levels: 1.5 nats equals the SC cap, so a saturated social-communication domain alone is screen-positive, and 1.8 nats lies just under the weight of a single certain stereotypy (feature~7, $w_7 = 1.838$), so high risk requires restricted or repetitive evidence. The decision is therefore a fixed function of the saved table~\cite{rudin2019stop}: any run can be re-scored, with weights or cut-offs re-tuned, without a model call, and each screen-positive total decomposes into the rows that produced it.

%%%%%%%%%%%%%%%%%%%%%%%%%%%%%%% Experiments %%%%%%%%%%%%%%%%%%%%%%%%%%%%%%%
\section{Experiments and Results}
\label{sec:results}

\subsection{Experimental Setup}
\label{subsec:exp_setup}

All models are used training-free (temperature $0$, \texttt{top\_p} $0.95$, \texttt{top\_k} $40$, \texttt{max\_output\_tokens} $65{,}536$). Seeds are the decoding seeds passed in the generation config; every configuration is run at three seed values and reported as mean~$\pm$~s.d. The backbone is Gemini 3.1 Pro~\cite{google2026gemini31} (\texttt{gemini-3.1-pro-preview}, high thinking) unless a table names another. Evaluation uses the 43-clip test set (12 ASD, 31 TD; Section~\ref{sec:dataset}); we report recall, specificity, F$_1$, and accuracy with ASD as the positive class, AUC where a configuration yields a continuous score, and count unparseable predictions as errors.

Three configurations recur through the section: (a) The \emph{zero-shot baseline} is a single VLM call over the video and the child's age with the glossary, the clinician persona, the four-step scaffold, and the structured output schema, without exemplars or age norms; its label is the VLM's own verdict. (b) The \emph{two-stage split} separates perception from judgment as in Section~\ref{sec:method}: Stage~1 writes the event table with the multi-turn exemplars, and the text-only Stage~2 applies the age calibration and returns a verdict; it has neither the grounded perception constraints nor the scorer, and its label is the LLM verdict of Stage~2. (c) The \emph{full pipeline} adds the grounded perception constraints to Stage~1 and replaces the LLM verdict with the deterministic scorer, intermediate and high risk both counting as screen-positive.

\subsection{Classification Performance}
\label{subsec:performance}

\begin{figure}[t]
\centering
\includegraphics[width=\textwidth]{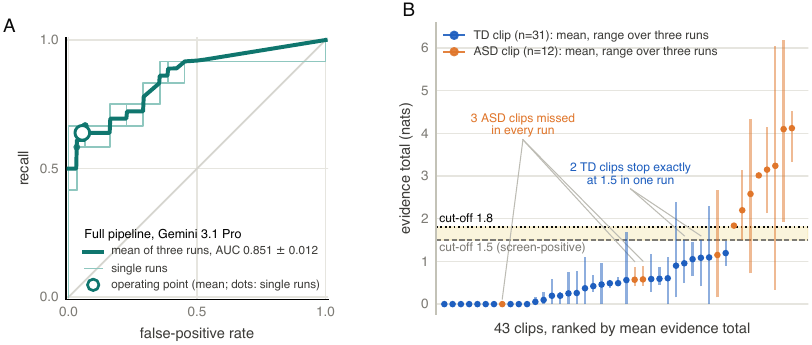}
\caption{Classification performance of the full pipeline of our method. \textbf{A}: ROC curve of the full pipeline, mean over the three runs (bold) and single runs (thin), with its operating point (large marker: mean; small markers: single runs). \textbf{B}: evidence total of every clip under the full pipeline, mean over the three runs (dot) and range (bar). Annotations mark the three ASD clips missed in every run and the two TD clips whose total stops exactly at 1.5 in one run.}
\label{fig:main_result}
\end{figure}

On the 43-clip cohort the full pipeline attains an AUC of 0.851~$\pm$~0.012 over the scorer's evidence total, with 86.0\%~$\pm$~0.0\% accuracy, 94.6\%~$\pm$~1.9\% specificity, 63.9\%~$\pm$~4.8\% recall, and F$_1$ of 71.8~$\pm$~1.6 across three runs, committing one or two false positives per run (Fig.~\ref{fig:main_result}A). At the two-stage split's false-positive rate of 8.6\% the curve gives 63.9\% recall against its 41.7\%; the zero-shot baseline's operating point (86.1\% recall at a false-positive rate of 37.6\%) lies on the curve. An age-only logistic baseline fitted in-sample reaches AUC 0.706 (older children skew ASD here).

Fig.~\ref{fig:main_result}B shows the margin behind these numbers: each clip's evidence total under the full pipeline, averaged over the three runs and with its run-to-run range, against the scorer's two fixed cut-offs. Ten of the 31 TD clips score zero in every run, 26 never reach the screen-positive cut-off, and the five that cross it do so in one run each, so no TD clip is flagged in more than one run. Two of the five stop exactly at the cut-off, at the social-communication cap of 1.5 nats with no repetitive-behavior evidence; the other three exceed it once, by 0.2 to 0.9 nats. Of the 12 ASD clips, five exceed the high-risk cut-off in every run, and three stay below the screen-positive cut-off in every run with near-empty event tables, which holds recall at or below 75.0\%. The run-to-run range of the total has a median of 0.38 nats for TD clips and 1.41 nats for ASD clips, and eight clips cross the screen-positive cut-off in some runs only. The cap and cut-offs were set on an earlier run over this cohort; re-selecting the cap and the cut-off by leave-one-out lowers pooled recall to 58--61\% and specificity to 91--95\%, and the cap, equal to the cut-off, cannot change a screen-positive decision (Appendix~\ref{subsec:scorer_twostage}).

\begin{table}[t]
\centering
\caption{Cumulative ablation of the full pipeline on Gemini 3.1 Pro (temperature $0$, three seeds each, mean~$\pm$~s.d.; positive class ASD). Each row adds one module to the row above. \emph{Verdict} names the source of the label: VLM (single call), LLM (the text-only Stage~2), or scorer.}
\label{tab:ladder}
\footnotesize
\setlength{\tabcolsep}{5pt}
\begin{tabular}{llcccc}
\toprule
\textbf{Configuration} & \textbf{Verdict} & \textbf{Acc (\%)} & \textbf{Recall (\%)} & \textbf{Spec (\%)} & \textbf{F$_1$ (\%)} \\
\midrule
Zero-shot baseline                       & VLM    & 69.0 $\pm$ 5.4 & 86.1 $\pm$ 4.8 & 62.4 $\pm$ 6.7 & 60.9 $\pm$ 4.8 \\
+ Few-shot exemplars                     & VLM    & 66.7 $\pm$ 1.3 & 55.6 $\pm$ 4.8 & 71.0 $\pm$ 0.0 & 48.1 $\pm$ 3.2 \\
+ Age calibration                        & VLM    & 65.9 $\pm$ 2.7 & 91.7 $\pm$ 0.0 & 55.9 $\pm$ 3.7 & 60.0 $\pm$ 1.9 \\
+ Two-stage split                        & LLM    & 77.5 $\pm$ 3.6 & 41.7 $\pm$ 8.3 & 91.4 $\pm$ 1.9 & 50.7 $\pm$ 9.0 \\
+ Grounded perception constraints        & LLM    & 81.4 $\pm$ 2.3 & 47.2 $\pm$ 4.8 & 94.6 $\pm$ 4.9 & 58.7 $\pm$ 1.4 \\
\textbf{+ Deterministic scorer (our full pipeline)} & scorer & \textbf{86.0 $\pm$ 0.0} & 63.9 $\pm$ 4.8 & \textbf{94.6 $\pm$ 1.9} & \textbf{71.8 $\pm$ 1.6} \\
\bottomrule
\end{tabular}
\end{table}

\subsection{Ablation Study}
\label{subsec:ladder}

The full pipeline is the endpoint of five cumulative additions to the zero-shot baseline, re-run on the same backbone (Table~\ref{tab:ladder}). Accuracy moves from 69.0\% at zero-shot to 77.5\% with the two-stage split, 81.4\% with grounded perception constraints, and 86.0\% once the deterministic scorer reads the same runs; over the last three rows recall rises from 41.7\% to 47.2\% to 63.9\% and F$_1$ from 50.7 to 58.7 to 71.8 while specificity stays at 91.4--94.6\%. The two-stage split raises accuracy by 11.6 points, mainly through specificity (55.9\% $\rightarrow$ 91.4\%) at a cost in recall (91.7\% $\rightarrow$ 41.7\%). The prompt-level additions shift the operating point without improving it: accuracy stays at 65.9--69.0\% over the first three rows, and the exemplars, added to an otherwise byte-identical prompt, trade 30.6 points of recall for 8.6 of specificity. The architectural additions raise specificity from 62.4\% at zero-shot to 91.4\% with the two-stage split and 94.6\% with the constraints, on the same backbone.

\subsection{Stable Correctness Across Repeated Runs}
\label{subsubsec:stability}

\begin{figure}[t]
\centering
\includegraphics[width=0.7\textwidth]{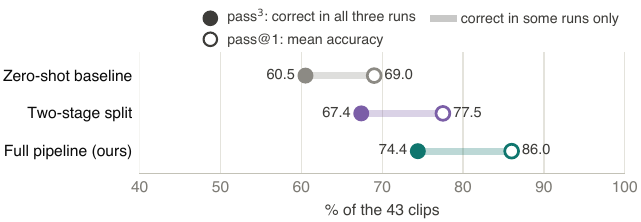}
\caption{Stable correctness of the three configurations of Section~\ref{subsec:exp_setup}. For each configuration, pass@1 and pass$^{3}$ as defined in Section~\ref{subsubsec:stability}; the shaded segment is the share of clips correct in some runs only.}
\label{fig:stable_correctness}
\end{figure}

Reasoning models sample an internal trace before the final answer, so a VLM verdict is stochastic even at temperature $0$: the single-call baseline (Gemini 2.5 Pro, five repeated runs of the identical request) averages 73.0\%~$\pm$~3.5\% accuracy with a best-to-worst spread of 7.0 points (per-run detail in Table~\ref{stab:stability}).
The instability is not confined to one configuration. With the cross-seed flip rate defined as the fraction of the 43 clips whose label differs across a configuration's repeated runs, the eight configuration--decision-rule combinations measured on Gemini 3.1 Pro span 16.3\% to 37.2\% (median 19.8\%; per-configuration rates in Table~\ref{stab:flips}). The metric counts binary verdicts near a decision boundary, where nondeterminism concentrates; the high run-to-run consistency reported for ordinal feature ratings on curated public clips~\cite{azizian2025mllmhomevideo} measures a different quantity. Fig.~\ref{fig:stable_correctness} reports two quantities per configuration: pass@1, the accuracy of a single run averaged over the three runs, and pass$^{3}$, the share of clips labeled correctly in all three runs~\cite{yao2024taubench}. pass$^{3}$ rises from 60.5\% at zero-shot to 67.4\% with the two-stage split and 74.4\% for the full pipeline, 26, 29, and 32 of the 43 clips; the gap to pass@1 stays between 8.5 and 11.6 points in all three, so a single run overstates what each configuration delivers in every run. Nine TD clips are flagged in every zero-shot run and none from the two-stage split onward, while five ASD clips are missed in every run of the two-stage split, with or without the constraints, and three by the full pipeline. The scorer recovers two of the five: their event tables carry at least 1.5 nats of evidence in each run while the LLM verdict reads TD in each run. The three clips missed in every run have near-empty event tables, so the remaining variation and the remaining misses both enter through the event tables.

%%%%%%%%%%%%%%%%%%%%%%%%%%%%%%% Discussion %%%%%%%%%%%%%%%%%%%%%%%%%%%%%%%

\section{Discussion and Conclusion}
\label{sec:discussion}
\label{sec:conclusion}

\looseness=-1 The operating point is consistent with confirmatory triage in an already-referred population: at this cohort's 27.9\% ASD prevalence, 94.6\% specificity gives precision 0.80 (8 of 10 flagged clips in one run), which would fall sharply at general-population prevalence, and 63.9\% recall means a negative result cannot exclude ASD. In the ablation, grounded perception constraints read through the deterministic scorer carry the gain; the three ASD clips missed in every run set the current recall ceiling at 75.0\%.

\looseness=-1 The deterministic decision layer changes what a reviewer receives. Each flagged row carries timestamps, a glossary label, and a rationale, and the verdict follows from those rows by arithmetic, so a disagreement can be traced to a specific event. Labels flipped across repeated runs in every configuration tested at temperature~0 (Section~\ref{subsubsec:stability}); the scorer removes the decision rule from that variance, and the clips labeled correctly in every run rise from 26 to 32 of 43 along the ablation. The scorer's gain over the LLM verdict holds when its cap and cut-off are re-selected by leave-one-out: pooled recall is 58--61\% at 91--95\% specificity (Appendix~\ref{subsec:scorer_twostage}), against 47.2\% at 94.6\% on the same tables (Table~\ref{tab:ladder}), and the ranking beats age alone (Section~\ref{subsec:performance}). The cut-off is also a parameter a site can set; raising it to the high-risk cut-off is the rule-in end of the dial (Appendix~\ref{subsec:scorer_twostage}). The zero-shot operating point, 86.1\% recall at a false-positive rate of 37.6\%, lies on the pipeline's curve (Fig.~\ref{fig:main_result}A), so a site that needs a rule-out screen can lower the cut-off on the same saved tables and read off the specificity it gives up. A model verdict offers no such dial, and in the ablation the prompt-level additions move recall by 30 to 36 points in either direction (Table~\ref{tab:ladder}).

\label{subsec:future}
\looseness=-1 The study includes 43 clips from one site, with only 12 ASD cases, and the DeLong interval ($[0.72, 1.00]$) is wide enough to admit a mediocre screen. Labels are clip-level expert ratings; one clip per child under-samples behavior; the scorer's cap and cut-offs were set on this cohort and cannot be validated on it; Stage~2's age norms are coarse; and nothing here speaks to ages, languages, or cultures outside the preschool, single-country cohort, so fairness across sex, language, and culture is untested. A multi-seed decomposition of the stages would sharpen the reliability account (Section~\ref{subsubsec:stability}). The main remaining bottleneck is perception: evidence missed in Stage~1 cannot be recovered downstream, by the scorer or by a lower cut-off. The three clips missed in every run are the concrete target. The next steps are better grounded behavioral observation, including prosody and caregiver speech for the under-detected reciprocity features, and external validation of the evidence layer at a second site and in prospective studies in which clinicians act on the tables\label{marker:contentend}.

%%%%%%%%%%%%%%%%%%%%%%%%%%%%%%% Acknowledgments (hidden in anonymous mode) %%%%%%%%%%%%%%%%%%%%%%%%%%%%%%%
\begin{ack}
This material is based upon work supported by the UAlbany Start-up funds for Prof. Xin Li. The funders had no role in study design, data collection and analysis, decision to publish, or preparation of the manuscript.
The authors declare that the research was conducted in the absence of any commercial or financial relationships that could be construed as a potential conflict of interest.

\textbf{Author contributions.} W.L. and M.R. are co-first authors and contributed equally to this work. C.H. contributes to the data collection and analysis of this project. S.W. and X.L. are responsible for securing the funding and supervising the project.
\end{ack}

\bibliographystyle{plainnat}
\bibliography{Reference}

%%%%%%%%%%%%%%%%%%%%%%%%%%%%%%% Appendix (former Supplementary Material) %%%%%%%%%%%%%%%%%%%%%%%%%%%%%%%
\clearpage
\appendix
\setcounter{table}{0}
\setcounter{figure}{0}
\renewcommand{\thetable}{A\arabic{table}}
\renewcommand{\thefigure}{A\arabic{figure}}
\renewcommand{\theHtable}{A\arabic{table}}   % keep hyperref anchors unique after the counter reset
\renewcommand{\theHfigure}{A\arabic{figure}}

\section*{Appendix}

\section{Cohort Composition}
\label{app:cohort}

\begin{table}[h]
\centering
\caption{Composition of the home-video corpus. Ages are the children's ages at recording. The four demonstration clips supply the in-context exemplars (Section~\ref{subsec:knowledge}) and are excluded from every reported metric.}
\label{stab:cohort}
\footnotesize
\setlength{\tabcolsep}{6pt}
\begin{tabular}{lccc}
\toprule
\textbf{Group} & \textbf{N} & \textbf{Age, mean $\pm$ s.d.\ (months)} & \textbf{Age range (months)} \\
\midrule
ASD (evaluation)                       & 12 & 40.9 $\pm$ 6.6 & 30--48 \\
TD (evaluation)                        & 31 & 36.6 $\pm$ 6.1 & 26--47 \\
\midrule
Demonstration clips (2 ASD, 2 TD)      & 4  & 35.8 $\pm$ 5.2 & 32--43 \\
\bottomrule
\end{tabular}
\end{table}

\section{Scorer: Derivation and Design Rationale}
\label{app:scorer_theory}
\textbf{Derivation.} Let $y\in\{\text{ASD},\text{TD}\}$ and let $E$ be the set of glossary features observed in a clip. Under conditional independence of the features given $y$, Bayes' rule gives
\[
\ln\frac{P(\text{ASD}\mid E)}{P(\text{TD}\mid E)} = \ln\frac{P(\text{ASD})}{P(\text{TD})} + {\textstyle\sum_{f\in E}}\ln\frac{\mathrm{sens}_f}{1-\mathrm{spec}_f} + {\textstyle\sum_{f\notin E}}\ln\frac{1-\mathrm{sens}_f}{\mathrm{spec}_f}.
\]
The scorer keeps only the second term, the sum of positive log-likelihood ratios $w_f=\ln\mathrm{LR}^{+}_f$ over observed features~\cite{good1950weighing,jaeschke1994lr}. The third term is dropped: in a 3--6 minute unprompted segment a behavior may never have the opportunity to occur, so absence is treated as uninformative and the total is a lower bound on the evidence for ASD. Bayesian combination of evidence is the oldest formalism of computer-aided diagnosis~\cite{ledley1959reasoning}; the additive weights-of-evidence form used here follows~\cite{spiegelhalter1984weights}.

\textbf{Confidence.} Events coded \texttt{possible} (occluded, brief, or downgraded by age calibration) contribute a quarter of their weight, a fixed discount for uncertain observations; each feature counts once, at its highest observed confidence, so repeated events cannot accumulate.

\textbf{Domain saturation.} Social-communication features co-occur because they express one underlying deficit, so their conditional independence fails and summing their weights double-counts; the social-communication total is therefore capped at $\kappa_{\mathrm{SC}} = 1.5$ nats, about one strong deficit ($\mathrm{LR}^{+}\approx 4.5$). Restricted and repetitive behaviors are near-independent of each other and are summed in full.

\textbf{Provenance.} The eleven sensitivity/specificity pairs (Table~\ref{stab:weights}) are author estimates, fixed before any reported run and not fitted to this cohort. One pair is taken directly from a published diagnostic study; feature~6 uses the symbolic play test's reported 0.72/0.62 at its optimal cut-off, estimated against a developmental-language-disorder comparison group rather than typically developing children~\cite{chang2023symbolicplay}. The remaining pairs were set from published effect sizes, prevalence contrasts, and classifier accuracies for the corresponding behavior; feature~7, for instance, from a prevalence contrast of 96--100\% (autism) against 25--63\% (typically developing, assessed up to 30 months) for at least one restricted or repetitive behavior in brief semi-structured observation~\cite{kim2010rrbtoddlers}. Feature~4 has no single anchoring study. The Basis column names the evidence consulted for each pair. Insistence on sameness (feature~8) is the least secure pair: it is uncommon before age three and did not separate groups in a toddler cohort~\cite{richler2007rrb}. The cap and the two cut-offs (low risk $<1.5$, intermediate $1.5$--$1.8$, high $\geq 1.8$ nats) are cohort-derived constants: they were set on an earlier run over this cohort, and Appendix~\ref{subsec:scorer_twostage} estimates the optimism of the cap and the screen-positive cut-off by leave-one-out re-selection.

\begin{table}[h]\centering
\caption{Per-feature weights of the deterministic scorer. Sensitivity and specificity are the assumed discriminative strength of the feature when present; $w_f=\ln(\mathrm{sens}_f/(1-\mathrm{spec}_f))$ in nats; the basis column names the type of evidence consulted for each pair.}
\label{stab:weights}
\footnotesize \setlength{\tabcolsep}{4pt}
\begin{tabular}{@{} c >{\RaggedRight}p{3.4cm} c c c >{\RaggedRight\arraybackslash}p{4.9cm} @{}}
\toprule
\textbf{\#} & \textbf{Feature} & \textbf{Domain} & \textbf{sens / spec} & $w_f$ & \textbf{Basis} \\
\midrule
1 & Unusual / avoidant eye contact & SC & 0.85 / 0.80 & 1.447 & eye-tracking meta-analysis~\cite{tecar2025eyetracking} \\
2 & Limited facial expressions \& social smiling & SC & 0.78 / 0.70 & 0.956 & high-risk sibling study; specificity limited~\cite{nichols2014socialsmiling} \\
3 & Reduced gestures \& joint attention & SC & 0.87 / 0.82 & 1.576 & joint-attention discriminability~\cite{dawson2004socialattention}; early joint-attention deficits predict 18-month screen risk~\cite{montagut2022jointattention} \\
4 & Preference for solitude / limited peer interest & SC & 0.80 / 0.79 & 1.338 & author estimate; multifactorial \\
5 & Impaired social reciprocity & SC & 0.88 / 0.84 & 1.705 & response-to-name marker studies~\cite{zhu2023responsetoname,miller2017responsetoname} \\
6 & Lack of imaginative / pretend play & SC & 0.72 / 0.62 & 0.639 & published pair, symbolic play test, ASD vs.\ language disorder~\cite{chang2023symbolicplay} \\
7 & Stereotyped motor movements & RRB & 0.88 / 0.86 & 1.838 & RRB prevalence contrast, ASD vs.\ TD~\cite{kim2010rrbtoddlers} \\
8 & Inflexible adherence to routines & RRB & 0.87 / 0.85 & 1.758 & DSM-5 criterion-set specificity~\cite{frazier2012dsm5validation,barton2013dsm5toddlers} \\
9 & Highly restricted / fixated interests & RRB & 0.83 / 0.82 & 1.528 & unusual-content specificity~\cite{burrows2021interests}; repetitive-interest prevalence contrast~\cite{kim2010rrbtoddlers} \\
10 & Echolalia \& atypical speech prosody & RRB & 0.81 / 0.75 & 1.176 & acoustic meta-analysis, ASD vs.\ TD~\cite{fusaroli2017voice}; prosody tracks severity within ASD~\cite{godel2023prosody} \\
11 & Sensory hyper-/hypo-reactivity \& unusual sensory interests & RRB & 0.85 / 0.84 & 1.670 & sensory-symptom prevalence, $\sim$90\% ASD vs.\ 5--15\% non-ASD, as reviewed in~\cite{riboldi2024sensory} \\
\bottomrule
\end{tabular}\end{table}

\subsection{Leave-one-out re-selection of the cap and the cut-offs}
\label{subsec:scorer_twostage}

The cap and the cut-offs were fixed on an earlier run over this cohort, before the three reported runs: the screen-positive cut-off at the social-communication cap (1.5 nats) and the high-risk cut-off at 1.8 nats, just under the weight of a single certain stereotypy ($w_7 = 1.838$). We quantify the optimism of this choice on the saved tables, without a model call, in three steps (Table~\ref{stab:loo}). 1) Re-selecting a single cut-off by the Youden index~\cite{youden1950} at the fixed cap returns 1.838 in 42 of 43 folds on the first run, 1.5 in 42 of 43 folds on the third, and alternates between 1.838 and 0.33 on the second: with 12 ASD clips the Youden point is not stable across runs, and the choice between the two levels turns on four clip-runs, two TD clips that stop exactly at the cap in one run each, one TD clip at 1.69 nats, and one ASD clip that reaches the cap in one run. Counting only high risk as screen-positive gives pooled recall 61.1\% and specificity 97.8\% over the three runs, against 63.9\% and 94.6\% at the deployed cut-off. 2) Re-selecting the cap (0.5 nats to no cap) and the cut-off jointly in every leave-one-clip-out fold, pooled over the three runs, gives recall 58.3--61.1\% and specificity 91.4--94.6\% under three selection criteria; the selected cap ranges from 0.5 to 1.75 nats across folds and criteria. 3) Because the screen-positive cut-off equals the cap, the cap cannot change a screen-positive decision; removing it moves two TD clip-runs from intermediate to high risk and the AUC by at most 0.008 (0.863 to 0.855, 0.839 to 0.839, 0.851 to 0.853); the AUC of the first run is 0.863 with DeLong 95\% CI~\cite{delong1988} $[0.72, 1.00]$.

\begin{table}[h]
\centering
\caption{Optimism of the fixed decision rule. Confusion counts, recall, specificity, and accuracy (\%) are pooled over the 43 clips of the three runs (129 clip-runs). The leave-one-out rows re-select the cap and the cut-off on 42 clips in every fold, under the named criterion, and apply them to the held-out clip.}
\label{stab:loo}
\footnotesize
\setlength{\tabcolsep}{5pt}
\begin{tabular}{llcccc}
\toprule
\textbf{Rule} & \textbf{Selection} & \textbf{TP/FN/TN/FP} & \textbf{Recall} & \textbf{Spec.} & \textbf{Acc.} \\
\midrule
Deployed rule (1.5) & set on an earlier run & 23/13/88/5 & 63.9 & 94.6 & 86.0 \\
High risk only (1.8) & set on an earlier run & 22/14/91/2 & 61.1 & 97.8 & 87.6 \\
Re-selected by leave-one-out & Youden index & 21/15/86/7 & 58.3 & 92.5 & 82.9 \\
 & maximum F$_1$ & 21/15/88/5 & 58.3 & 94.6 & 84.5 \\
 & recall first, spec.\ $\geq$ 90\% & 22/14/85/8 & 61.1 & 91.4 & 82.9 \\
\bottomrule
\end{tabular}
\end{table}

\section{Cross-Seed Flip Rates by Configuration}
\label{app:flips}

Table~\ref{stab:flips} lists the configurations behind the flip range quoted in Section~\ref{subsubsec:stability}.

\begin{table}[h]
\centering
\caption{Cross-seed flip rates on Gemini 3.1 Pro (three repeated runs at temperature $0$). \emph{Flip} is the fraction of the 43 clips whose label differs across the runs; \emph{Verdict} as in Table~\ref{tab:ladder}.}
\label{stab:flips}
\footnotesize
\setlength{\tabcolsep}{6pt}
\begin{tabular}{llc}
\toprule
\textbf{Configuration} & \textbf{Verdict} & \textbf{Flip (\%)} \\
\midrule
Zero-shot baseline                                     & VLM    & 16.3 \\
+ Few-shot exemplars                                   & VLM    & 18.6 \\
Single call, multi-turn exemplars, no age calibration  & VLM    & 37.2 \\
+ Age calibration                                      & VLM    & 23.3 \\
+ Two-stage split                                      & LLM    & 20.9 \\
Two-stage split, read by the scorer                    & scorer & 27.9 \\
+ Grounded perception constraints                      & LLM    & 16.3 \\
Full pipeline                                          & scorer & 18.6 \\
\bottomrule
\end{tabular}
\end{table}

\section{Run-to-run Stability of the Single-Call Baseline}
\label{app:singlecall}

Per-run detail of the five-run experiment summarized in Section~\ref{subsubsec:stability}: the single-call baseline (Gemini 2.5 Pro, full prompt and four in-context videos, VLM verdict) repeated five times as the identical request at temperature $0$ with a shared context cache. In three runs the API returned an empty response for the same clip; it counts as an error, as in Section~\ref{subsec:exp_setup}.

\begin{table}[h]
\centering
\caption{Run-to-run stability of the single-call baseline (Gemini 2.5 Pro, full prompt, four in-context videos, VLM verdict) over five repeated runs of the identical request at temperature $0$ (43 clips; an empty API response counts as an error).}
\label{stab:stability}
\footnotesize
\setlength{\tabcolsep}{6pt}
\begin{tabular}{lcccc}
\toprule
\textbf{Run} & \textbf{Accuracy (\%)} & \textbf{Recall (\%)} & \textbf{Specificity (\%)} & \textbf{F$_1$ (\%)} \\
\midrule
run 1 & 72.1 & 58.3 & 77.4 & 53.8 \\
run 2 & 69.8 & 58.3 & 74.2 & 51.9 \\
run 3 & 76.7 & 75.0 & 77.4 & 64.3 \\
run 4 & 69.8 & 66.7 & 71.0 & 55.2 \\
run 5 & 76.7 & 75.0 & 77.4 & 64.3 \\
\midrule
\textbf{mean $\pm$ s.d.} & \textbf{73.0 $\pm$ 3.5} & \textbf{66.7 $\pm$ 8.3} & \textbf{75.5 $\pm$ 2.9} & \textbf{57.9 $\pm$ 6.0} \\
\bottomrule
\end{tabular}
\end{table}

\section{Assets: Glossary, Prompt Structure, Norms, and Demonstrations}
\label{app:prompts}

\subsection{Prompt structure}
\label{app:prompt_structure}

\begin{figure}[h]
\centering
\includegraphics[width=\textwidth]{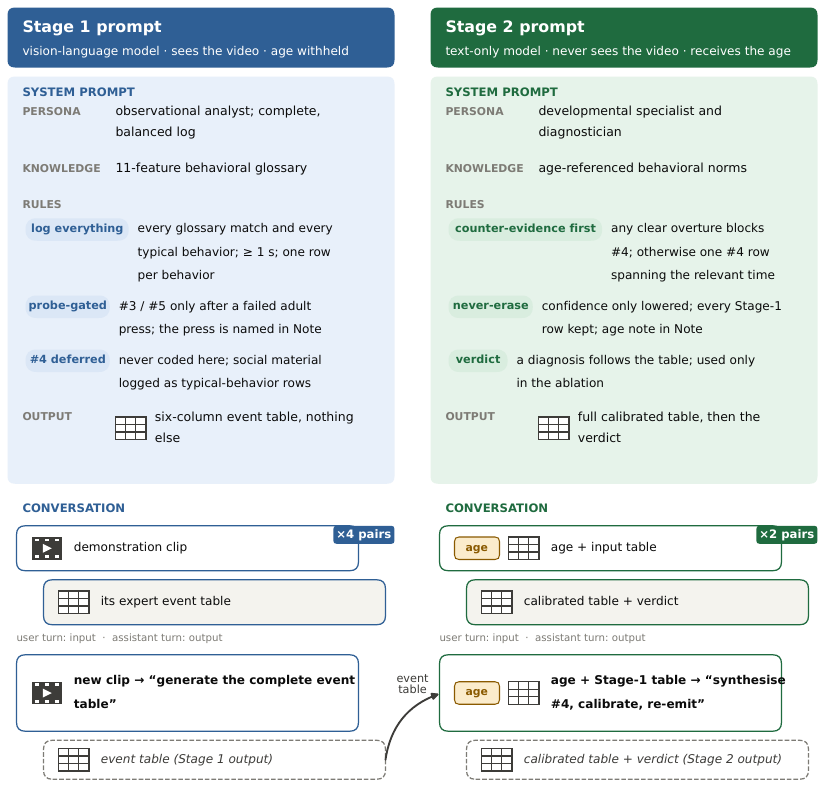}
\caption{Anatomy of the Stage~1 and Stage~2 prompts. Top: the system prompt of each stage (persona, knowledge, rules, output mandate); the knowledge blocks are the glossary of Table~\ref{tab:glossary} and the norms of Fig.~\ref{sfig:age_norms}. Bottom: the conversation the model sees, with demonstrations as prior user/assistant turn pairs (four corpus clips with their expert tables for Stage~1; the two hand-written exchanges of Fig.~\ref{sfig:demos} for Stage~2) and the final user turn; the dashed bubble is the stage's output, and Stage~1's event table becomes the content of Stage~2's final turn.}
\label{sfig:prompt_structure}
\end{figure}

Fig.~\ref{sfig:prompt_structure} summarizes the two prompts block by block; the rules are the prompts' own, shortened.

\subsection{Behavioral glossary}
\label{app:glossary}

The eleven features of Table~\ref{tab:glossary} condense a 22-behavior, nine-task home-video protocol that the clinical team designed in 2020 and mapped item by item to ADOS-2 algorithm codes and M-CHAT-R questions; we kept the behaviors codeable from unprompted free play.

\begin{table}[h]
  \centering
  \caption{The 11-item behavioral glossary: feature (with ADOS-2 / DSM-5 mapping where applicable), behavioral description, typical contexts, and an example. It constrains both expert annotation and model prompting.}
  \label{tab:glossary}
  \scriptsize
  \renewcommand{\arraystretch}{1.05}
  \setlength{\tabcolsep}{3pt}
  \begin{tabularx}{\linewidth}{@{} c >{\RaggedRight}p{2.6cm} >{\RaggedRight}X >{\RaggedRight}X >{\RaggedRight\arraybackslash}X @{}}
    \toprule
    \textbf{\#} & \textbf{Feature (merged diagnostic items)} & \textbf{Expanded behavioral description} & \textbf{Typical situations / contexts} & \textbf{Representative examples} \\
    \midrule
    1 & \textbf{Unusual / Avoidant Eye Contact} (ADOS-2 SA ``Unusual Eye Contact'', DSM-5 A2 component) & Limited, fleeting or sidelong gaze; may avoid looking at faces even while interacting. & Feeding, play, greetings, clinical exams. & \textit{Infant does not lock eyes during feeding; toddler continues playing when a visitor enters and does not look up despite name being called.} \\
    \midrule
    2 & \textbf{Limited Facial Expressions \& Social Smiling} & Flat or restricted affect; smiles and emotive expressions are infrequent, delayed, or context-inappropriate. & Peek-a-boo, praise moments, gift giving, celebrations. & \textit{Child watches peek-a-boo with neutral face; preschooler receives a gift with minimal facial change.} \\
    \midrule
    3 & \textbf{Reduced Gestures \& Joint Attention} & Pointing, showing, waving, and nodding are rare; may not coordinate gaze with gesture or follow others' pointing. & Requesting help, picture-book reading, playground pointing. & \textit{Child reaches toward a toy but does not point or look to a caregiver to share the request.} \\
    \midrule
    4 & \textbf{Preference for Solitude / Limited Peer Interest} & Appears content playing alone; seldom approaches peers or adults to share enjoyment or comfort. & Day-care free-play, playdates, family gatherings. & \textit{Child ignores a peer offering a toy and continues solitary play.} \\
    \midrule
    5 & \textbf{Impaired Social Reciprocity} & Reduced back-and-forth interaction; may not respond to name or may show limited verbal turn-taking. & Name-call tests, greetings, Q\&A exchanges. & \textit{Child does not orient when name is called after multiple attempts.} \\
    \midrule
    6 & \textbf{Lack of Imaginative / Pretend Play} & Symbolic play is absent or limited; play remains concrete, repetitive, or focused on object properties. & Doll/kitchen sets, toy-vehicle play, cooperative make-believe. & \textit{Child spins toy-car wheels and does not use the car in pretend scenarios.} \\
    \midrule
    7 & \textbf{Stereotyped Motor Movements} (DSM-5 B1; colloquially ``stimming'') & Stereotyped motor acts such as hand-flapping, rocking, or spinning; toe walking is noted separately as an associated motor sign. & Idle moments, excitement peaks, transitions. & \textit{Child flaps hands repeatedly while watching a preferred stimulus.} \\
    \midrule
    8 & \textbf{Inflexible Adherence to Routines} & Distress with small changes to routines, objects, routes, or expected sequences. & Morning routines, classroom agendas, travel routes. & \textit{Child refuses a meal when a preferred item or arrangement is changed.} \\
    \midrule
    9 & \textbf{Highly Restricted / Fixated Interests} & Narrow interests of unusually high intensity that dominate play or attention. & Free-choice play, conversations, preferred media. & \textit{Child focuses on one object category and rejects alternative toys.} \\
    \midrule
    10 & \textbf{Echolalia \& Atypical Speech Prosody} & Repetition of words or phrases; scripted speech; flat, sing-song, or atypical intonation. & Adult questioning, solitary play scripting. & \textit{Child repeats a question verbatim rather than answering it.} \\
    \midrule
    11 & \textbf{Sensory Hyper-/ Hypo-reactivity \& Unusual Sensory Interests} & Over- or under-response to sensory input; unusual fascination with sensory aspects of objects. & Noise/lights, swings, mealtimes, medical visits. & \textit{Child covers ears and cries in response to common sounds; or repeatedly seeks spinning visual stimuli.} \\
    \bottomrule
  \end{tabularx}
\end{table}

\clearpage
\subsection{Age-referenced norms}
\label{app:norms}

\begin{figure}[h]
\centering
\includegraphics[width=\textwidth]{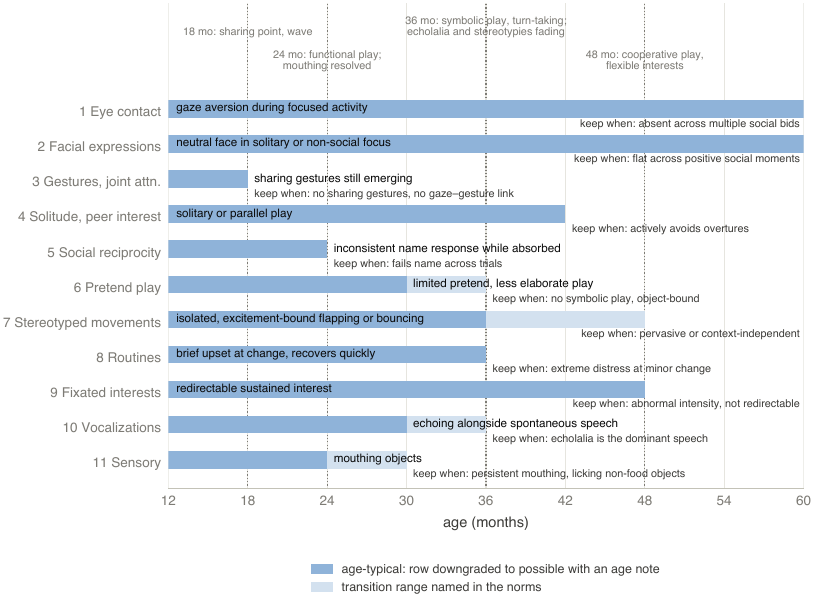}
\caption{The age-referenced norms of Stage~2, read from the norms text. For each glossary feature the bar marks the ages at which the named behavior is age-typical, so the row is downgraded to \texttt{possible} with an age note; the light segment is a transition range the norms name; the line under the bar is the condition under which the feature is kept. Dotted lines are the developmental anchors the norms list. Pervasiveness overrides the window at any age.}
\label{sfig:age_norms}
\end{figure}

For every feature row, Stage~2 checks the behavior against the norm for the child's age. A behavior that is age-typical keeps its feature number, drops to \texttt{possible}, and receives an age note; \texttt{certain} is reserved for behaviors that clearly exceed the age expectation in intensity, duration, or pervasiveness. Pervasiveness overrides the exemptions: the age-typical conditions apply only to occasional, isolated instances, and a behavior that recurs or dominates the clip stays \texttt{certain} even when each instance would be exemptible. Fig.~\ref{sfig:age_norms} places the eleven features on the age axis.

\subsection{Stage~2 demonstrations}
\label{app:prompt_stage2}

\begin{figure}[p]
\centering
\includegraphics[width=\textwidth]{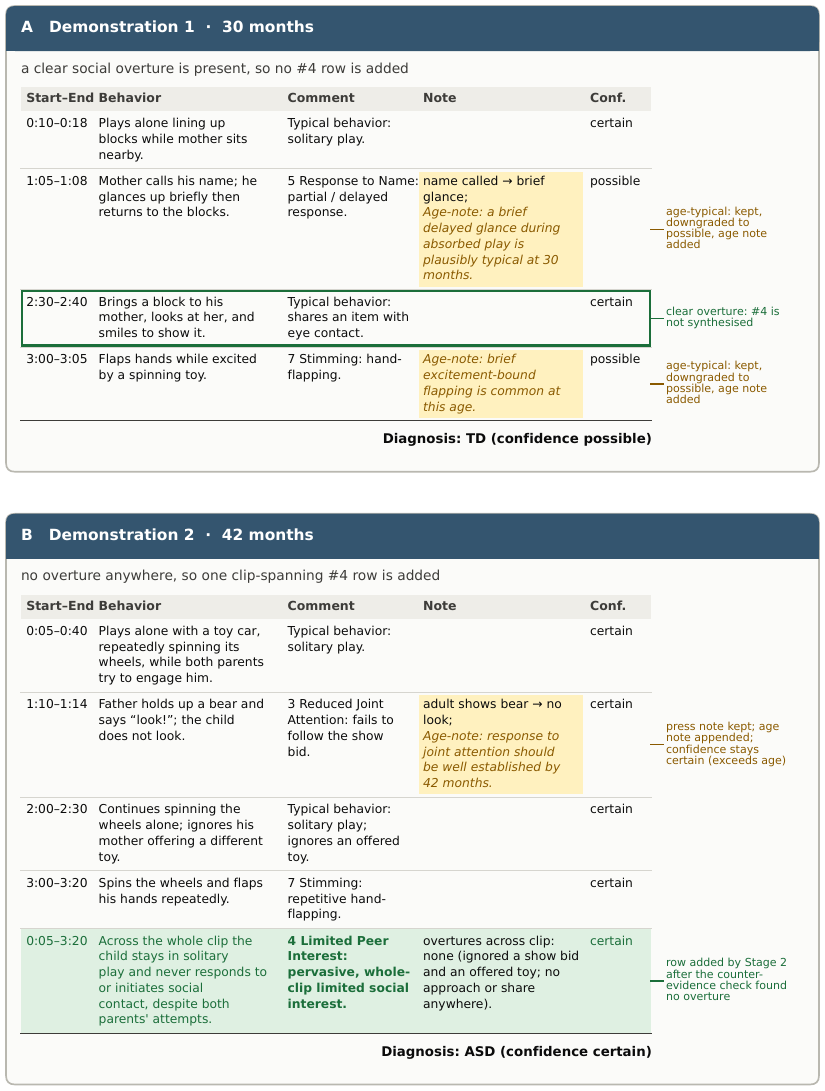}
\caption{The two hand-written Stage~2 demonstrations, drawn as the output tables (\textbf{A}: 30 months; \textbf{B}: 42 months). Yellow cells are the age notes Stage~2 wrote into the Note column; the green outline in A marks the social overture that blocks the pervasive feature \#4; the green row in B is the clip-spanning \#4 row Stage~2 added after the counter-evidence check found no overture. Feature numbers refer to Table~\ref{tab:glossary}.}
\label{sfig:demos}
\end{figure}

Two hand-written exchanges precede the clip's turn, each a user turn (the user prompt with the age and the input table) followed by an assistant turn (the output). Fig.~\ref{sfig:demos} shows the two outputs with Stage~2's changes marked; the input tables are identical except for those changes.

\end{document}